\documentclass[letterpaper, 10 pt, conference]{ieeeconf}
\IEEEoverridecommandlockouts    % to override locked commands

\usepackage{multirow}
\usepackage{lipsum}
\usepackage{amsmath}
\usepackage{amssymb}
\usepackage[linesnumbered,ruled,vlined]{algorithm2e}
\usepackage{graphicx}
\usepackage{booktabs}
\usepackage{graphicx}
\usepackage{tabularx}
\usepackage{xcolor}
\graphicspath{{./Figures/}}
\usepackage{subcaption}
\usepackage{siunitx}
\usepackage{censor}
\usepackage[T1]{fontenc}
\usepackage{tikz}

\newcommand\copyrighttext{%
\footnotesize \copyright 2026 IEEE. Personal use of this material is permitted. Permission from IEEE must be obtained for all other uses, in any current or future media, including reprinting/republishing this material for advertising or promotional purposes, creating new collective works, for resale or
redistribution to servers or lists, or reuse of any copyrighted component of this work in other works.}

\newcommand\copyrightnotice{%
\begin{tikzpicture}[remember picture,overlay]
\node[anchor=north,yshift=-1cm] at (current page.north) {\parbox{\dimexpr\textwidth-\fboxsep-\fboxrule\relax}{\centering \copyrighttext}};
\end{tikzpicture}%
}

\begin{document}
\title{\LARGE \bf Mission Performance: Automatic and Adaptive Race Pace Progression for Autonomous Racing }


% \author{\IEEEauthorblockN{
% \censor{Giovanni Lambertini\IEEEauthorrefmark{1}, 
% Matteo Pini\IEEEauthorrefmark{1}, 
% Nicola Musiu\IEEEauthorrefmark{1},
% Ayoub Raji\IEEEauthorrefmark{1}, 
% Francesco Iacovacci\IEEEauthorrefmark{1}, 
% Marko Bertogna\IEEEauthorrefmark{1}
% }}
% \IEEEauthorblockA{\IEEEauthorrefmark{1}\censor{University of Modena and Reggio Emilia, Modena, Italy\\ 
% giovanni.lambertini@unimore.it}}
% }

\author{
    \parbox{\textwidth}{%
        \centering
        Giovanni Lambertini, Matteo Pini, Nicola Musiu, Ayoub Raji, Francesco Iacovacci, and Marko Bertogna%
    }%
\thanks{All authors are with the Department of Physics, Informatics and Mathematics, University of Modena and Reggio Emilia, 41121 Modena, Italy.
{\tt\small \{firstname.lastname\}@unimore.it}}
}

\hyphenation{tem-por-ary}

%%%%%%%%%%%%%%%%%%%%%%%%%%%%%%%%%%%%%%%%%%%%%%%%%%%%%%%%%%%%%%%%%%
	
\maketitle
\copyrightnotice
\thispagestyle{empty}
\pagestyle{empty}

%%%%%%%%%%%%%%%%%%%%%%%%%%%%%%%%%%%%%%%%%%%%%%%%%%%%%%%%%%%%%%%%%%
\begin{abstract}
    In this paper, we describe the Mission Performance module implemented for a fully autonomous racing car to automatically manage the longitudinal, lateral, and combined performances, aiming to speedup the laptime progression while assuring safety. Motivated by the difficulty and risks of applying the real-time estimation of the grip to critical modules like the motion planner and controller, the Mission Performance guides these modules adapting their target performance instead of changing the vehicle model parameters. The module is formed by pre-defined progressions to warm up the tires at the beginning of a run. Then, the system continuously monitors safety and vehicle dynamics metrics on a per-sector basis to adaptively reduce, maintain, or increase the performance levels for each sector, progressively converging toward the maximum allowed value. The solution's effectiveness is demonstrated on the EAV-25, a fully autonomous Dallara Superformula, at the Yas Marina Circuit during the Abu Dhabi Autonomous Racing League (A2RL) Season 2.
\end{abstract}

%%%%%%%%%%%%%%%%%%%%%%%%%%%%%%%%%%%%%%%%%%%%%%%%%%%%%%%%%%%%%%%%%%
\section{Introduction}
\label{sec:introduction}
    In recent years, autonomous racing has gained increasing importance, proving to be a powerful tool for pushing the validation of perception, planning, and control algorithms to their physical limits \cite{betz_survey}. Competitions like Indy Autonomous Challenge (IAC) \cite{iac} and Abu Dhabi Autonomous Racing League (A2RL) \cite{a2rl} challenge university teams to compete with full-scale autonomous race cars at high speeds. In these events, vehicles are required to race fully autonomously, involving also multi-car interactions, from the moment they exit the pit lane until they return at the end of the run. This introduces a critical operational challenge: managing the vehicle's performance without human intervention. 
    
    In traditional motorsport, human drivers instinctively balance safety and performance, adapting in real-time to variable tires and track conditions. They can anticipate reduced grip due to cold tires, debris, or moisture. In autonomous systems, this adaptation must be coded. While the objective is to exploit the full friction potential, a mechanism to continuously validate vehicle stability is fundamental. Consequently, the race strategy requires an adaptive progression that balances safety and performance, aiming to approach the vehicle handling limits in the shortest possible time. This is achieved by leveraging real-time metrics to dynamically scale down performance whenever necessary.

    In this paper, we present the Mission Performance module of our autonomous racing stack. Its objective is to automate the management of lateral, longitudinal, and combined performance while preserving safety without any human intervention. The performances are increased every lap using a pre-defined progression to warm up the tires. At the same time, the system evaluates specific safety and vehicle dynamics metrics to assess the vehicle state in real-time, limiting performance as necessary to maintain vehicle stability. A tire temperature-based Lookup Table (LUT) is also exploited to adaptively update the performance in case of temperature drops.

    Related works on race strategies and the contribution of the paper are given in Section~\ref{sec:related_works}. An overview of the motion planning system influenced by the Mission Performance is presented in Section~\ref{sec:system}, as well as the analysis of a Grip Estimation module, whose outcome motivates the proposed solution. In Section~\ref{sec:designandimplementation}, the design and implementation of the Mission Performance is described. The results of the automatic and adaptive performance progression applied to the A2RL Grand Finale race at the Yas Marina Circuit, and a specific test in which the racecar had an oversteering condition, are presented in Section~\ref{sec:results}. Conclusions and future extensions are presented in Section~\ref{sec:conclusion}.

    \section{Related Works and Contribution}
    \label{sec:related_works}
    In traditional motorsport, race strategy is a mature field. The literature offers extensive solutions for minimizing lap times while adapting to energy/fuel consumption.
    Balerna et al. \cite{fuel_minimization} present low-level control strategies to minimize the fuel consumption in a high-performance hybrid electric power unit. 
    Van Kampen et al. \cite{endurance_thermal_constraints} present a strategy that maximizes race distance and explicitly models the thermodynamics of the battery and electric motors, preventing power cuts due to overheating for a fully electric endurance racing car. 
    Borsboom et al. \cite{lap_time_minimization_electric} propose a methodology to optimize a single lap time, exploiting the full potential of the powertrain and available grip.
    
    Extensive research has been dedicated to strategy optimization in Formula 1. Some of these works, such as \cite{time_optimal_control} and \cite{minimum_lap_time_control}, specifically focus on exploiting the hybrid power unit to minimize the single lap time. The optimal control policy manages the torque split and energy recovery to fully exploit the available battery resources over a single lap.  In \cite{adaptation_algorithms_powertrain}, online adaptation mechanisms are introduced, adjusting the energy deployment in real-time to compensate for vehicle or track changes.
    In \cite{equivalent_lap_time_minimization}, the authors propose a long-term race management strategy that balances fuel consumption against lap time reduction, optimizing the pace over the entire race distance.
    
    Regarding tire strategy, the literature predominantly focuses on long-term tire degradation and pit-stop planning. In \cite{optimal_tyre_management}, Wilhelm J. West et al. propose strategies to manage physical wear and thermal limits within a stint, and in \cite{optimizing_pit_stop_management_game_theory}, Felipe Aguad et al. apply game theory to optimize high-level pit-stop timing and compound selection against opponents.

    In general, the focus is on performance optimization, assuming a proficient human driver handles the vehicle dynamics, such as the vehicle's stability during the warm-up phase or under varying track and vehicle conditions. This assumption cannot always be guaranteed, particularly for a fully autonomous agent.

    Regarding the autonomous racing context, literature mainly focuses on lap time optimization.  
    Works such as \cite{learning_policies_automated_racing} and \cite{energy_management_autonomous} focus on optimizing single-lap time, employing learning policies and energy management to push the vehicle to its physical limits. Extending this scope to the full race distance, \cite{minimum_race_time_planning_strategy_autonomous} shifts the objective from the fastest lap to the minimum total race time. This framework optimizes energy deployment over multiple laps, managing long-term thermal and battery constraints to maximize the overall pace.

    Promising results in online model adaptation have been demonstrated in works such as \cite{learing_based_mpc_amz}, which employs Gaussian Processes to correct vehicle dynamics on a Formula Student prototype. However, this validation is limited to relatively low velocities. Applying such dynamic learning strategies to full-scale race cars operating at high speeds introduces severe safety risks, as aggressive model updates at the limit of handling can lead to instability, particularly on optimization-based algorithms.
   
    The cited works have the goal of minimizing the use of resources (energy, fuel) or of the lap time in general, with some degree of online adaptation. None of them explicitly focus on having an increasing and adaptive progression based on safety and vehicle stability. To the best of the authors' knowledge, this is the first work to address the management of a complete autonomous race, starting from the warm-up phase and adapting the performance to the track conditions in real time.

%%%%%%%%%%%%%%%%%%%%%%%%%%%%%%%%%%%%%%%%%%%%%%%%%%%%%%%%%%%%%%%%%%
\section{System Overview}
    \label{sec:system}
    The main objective of the Mission Performance is to manage autonomous racing events, especially targeting A2RL format. In these multi-agent competitions, up to six vehicles race simultaneously without any human intervention. All competitors share an identical hardware platform, including the sensor suite, tire pressure, and aerodynamic setup. The distinguishing factor between teams relies entirely on the software stack. Prior to the events, all cars utilize tire warmers to pre-heat the compounds. 
    On-track operations are managed by Race Control via a digital flag system. A \textit{PIT\_OUT} flag signals the start of the reconnaissance laps, with a 100~km/h speed limit for the leader. Once the group is formed and the lead car crosses the start/finish line, the \textit{GREEN} flag signals the start of the race to unlimited speed, authorizing autonomous overtakes. Furthermore, Race Control dynamically manages unforeseen track hazards. For example, the \textit{YELLOW} flag is sent to all the cars to limit the speed to 80 km/h in the event of an incident.

    \begin{figure}[t]
      \centering
      \vspace{6pt} % Adds a small margin above the figure
      \includegraphics[width=1.0\columnwidth]{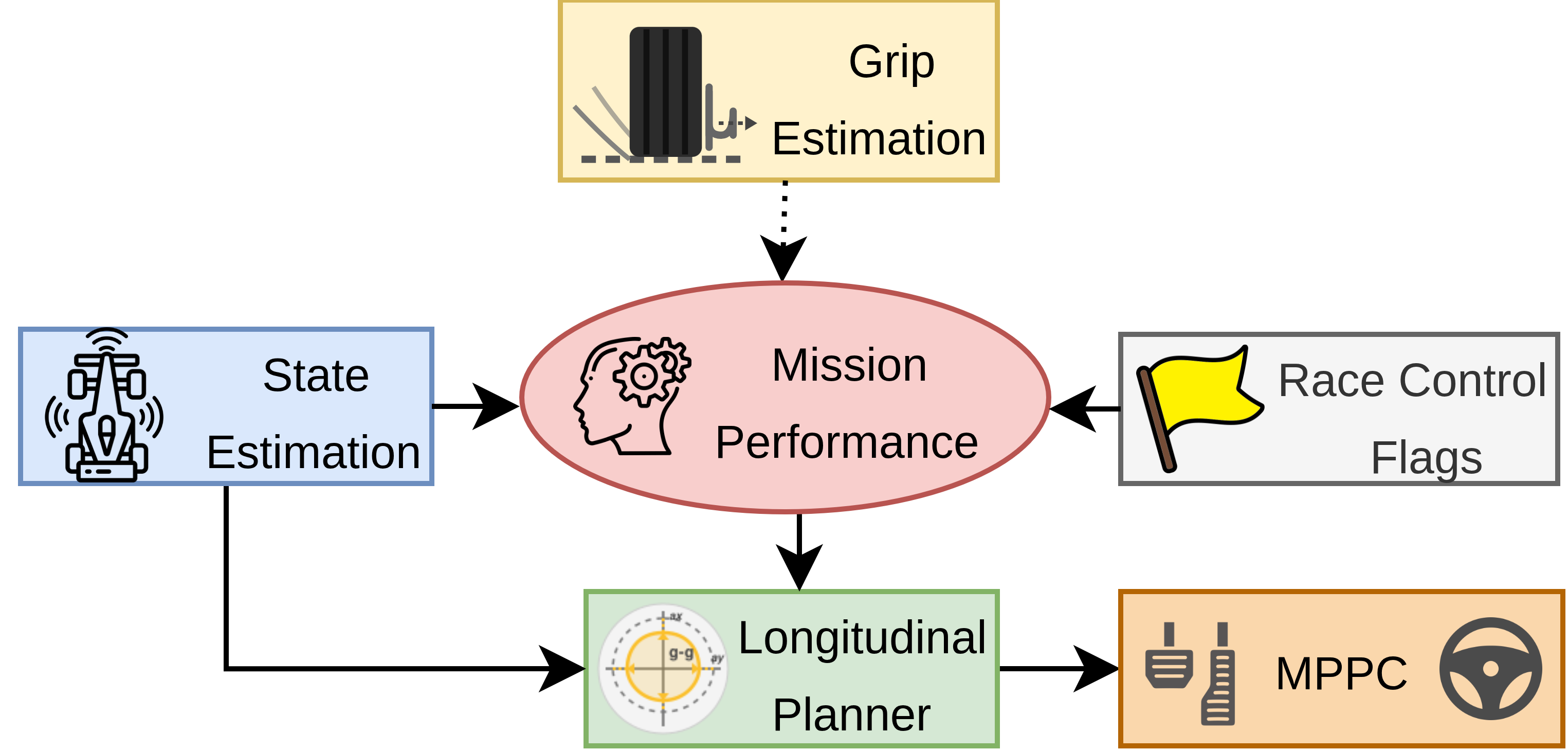}
      \caption{Architectural subset of the autonomous racing stack focused on the Mission Performance module and its interacting components.}
      \label{img:system_architecture}
    \end{figure}
    
    The following subsections briefly introduce the overall system architecture depicted in Figure \ref{img:system_architecture}. Implementation details about modules external to the Mission Performance, such as Grip Estimation, Planners, and Controller, are avoided, focusing only on the aspects that are influenced by the automatic performance progression and that motivate its application. Further details will be given in future work on the complete software stack.
 
\subsection{Grip Estimation}\label{sub_sec:grip}
% \begin{figure}[] 
%             \begin{subfigure}{\columnwidth}
%             \centering
%             \includegraphics[width=.9\linewidth, trim=0 0 0 0, clip]
%             {figures/grip_over_time.pdf}
%             \caption{Time evolution of the identified grip, starting 
%           from the initial guess. The lower plot reports the average tire 
%           temperature per axle.}
%             \label{fig:grip_ev}
%             \end{subfigure}
            
%             \vspace{5pt}
            
%             \begin{subfigure}{\columnwidth}
%                 \centering
%                 \includegraphics[width=1.\linewidth, trim=0 0 0 0, clip]
%                 {figures/pacejka_progression.pdf}
%                 \caption{Point cloud representing the experimental data.
%           The blue-gradient curves represent the identified tire model evolution over 
%           time, with darker shades corresponding to normalized later estimation steps.\cambiare{toglierei}}
%                 \label{fig:pacejka}
%             \end{subfigure}
            
%             \caption{ Results of the tire grip estimation and its dependency on tire’s operating conditions.}
%             \label{fig:grip_est}
% \end{figure}

\begin{figure}[b]
\centering
\includegraphics[width=1.\linewidth, trim=0 0 0 0, clip]{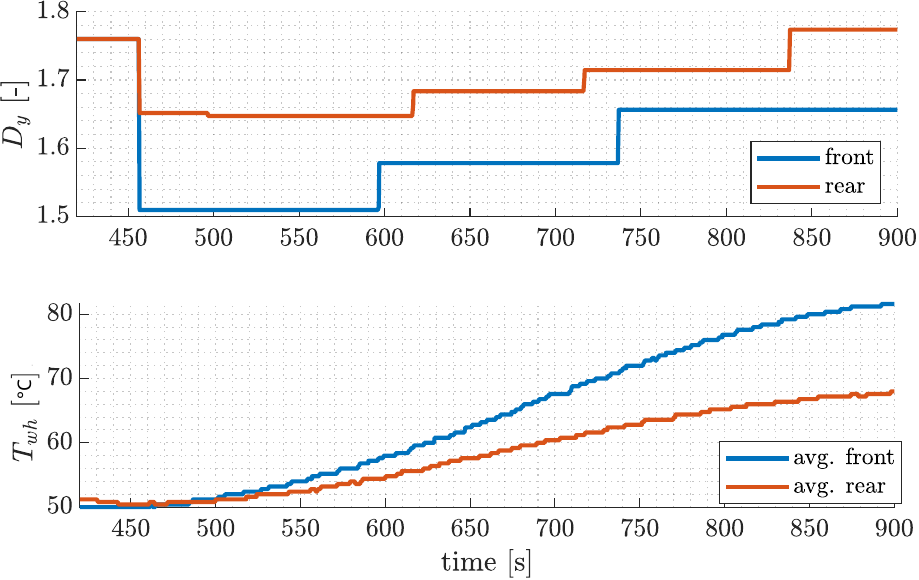}
\caption{
Time evolution of the identified lateral grip ($D_y$), starting from the initial guess. The lower plot reports the average tire temperature per axle ($T_{wh}$).
}
\label{fig:grip_ev}
\end{figure}

The grip estimation module performs online adaptation of the tire model and, consequently, the 
tire-road friction levels ($D_y$) across evolving tire conditions.
Given the vehicle parameters (mass, inertia, aerodynamic coefficients), 
tire slips ($\alpha$), lateral forces ($F_y$), and vertical loads ($F_z$) are estimated to reconstruct the experimental relationship $\frac{F_y}{F_z} = f({\scriptstyle \alpha})$ for the front and rear axles. The selected tire model is the Pacejka Magic Formula \cite{pacejka}. 
Its parameters are identified by solving a nonlinear least-squares problem
formulated using Ceres Solver \cite{ceres}.

An example of the estimation process during an experimental test is shown in Figure~\ref{fig:grip_ev}. The upper plot shows the evolution of the identified peak parameters $D_y$, which represent the maximum lateral friction coefficient at each axle. 
The lower plot reports the average axle tire temperature measured from the inner tire core (TPMS sensors).
In the lower plot, tire temperature emerges as the dominant factor affecting parameter variation: as temperature increases, the available lateral grip rises, consistently with the known behavior of racing tires \cite{farroni_temp}.

Starting from an initial guess of $D_{yf} = D_{yr} = 1.75$, the update mechanism is triggered when the estimated friction deviates by more than 5\% from the nominal value (around $t \approx 450$ s). 
The parameters then evolve progressively and converge to steady-state values of approximately 
$D_{yf} = 1.66$ and $D_{yr} = 1.76$.
Compared to the initial guess, the identified parameters indicate a more understeering vehicle balance, consistent with the high tire mileage during the test (over 250 km).

% \nic{
% Subfig.~\ref{fig:pacejka} reports the dataset collected during the test, consisting of approximately $N \approx 7800$ samples $(\alpha_i, D_{yi})$ per axle. 
% The black and orange point clouds represent the stored measurements, while the blue-gradient curves denote the identified axle characteristics, with color intensity increasing over time.
% Since the performed maneuvers involved nearly pure lateral dynamics, only negligible differences are observed between the two datasets.
% The shift coefficient $S_h$ was set to zero, as the vehicle setup is expected to
% be symmetric. Nevertheless, a slight asymmetry appears in the front axle data,
% causing the identified parameters to converge toward a trade-off that balances
% positive and negative slip angle behavior.
% }

\subsection{Mission Performance}\label{sub_sec:miss_perf}

\begin{figure}[b]
\centering
\includegraphics[width=1.\linewidth, trim=285 0 255 0, clip]{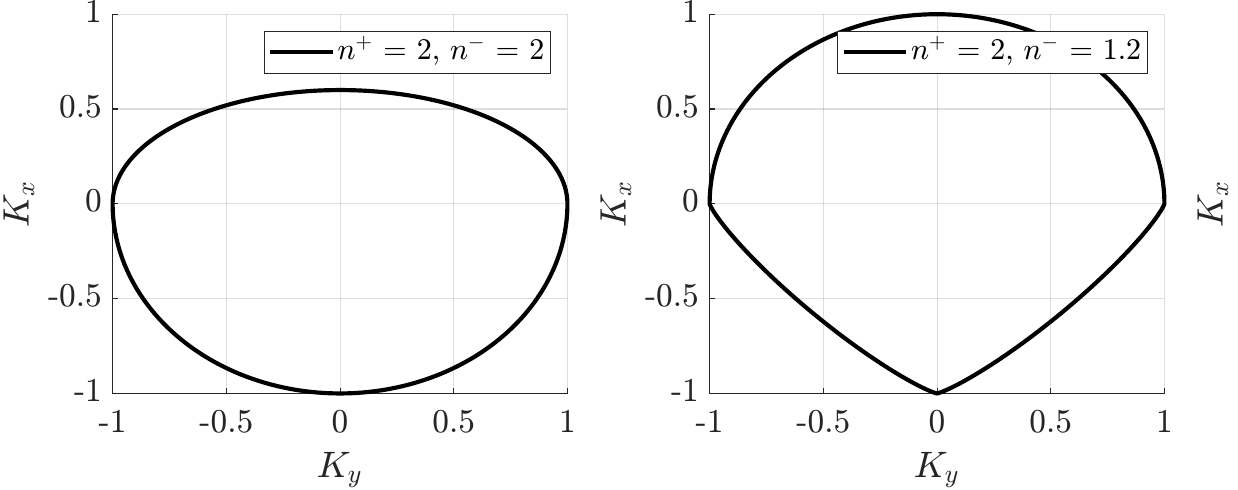}
\caption{Mission Performance under two operating conditions.
Left: reduced acceleration with full combined effect.
Right: full performance with reduced braking combined effect.}
% \caption{
% Four subplots illustrating different operating conditions of the mission planner.
% From left to right: full pure performance with no combined slip, reduced accel-
% eration performance with full combined effect, reduced braking combined effects,
% and full pure and combined performance.
% }
\label{img:GG}
\end{figure}
        
It defines high-level performance targets for the longitudinal planner (Sec.~\ref{sub_sec:l_plan}). It enables independent tuning of lateral performance ($K_y$), longitudinal acceleration ($K_{x+}$), longitudinal braking ($K_{x-}$), and combined tire utilization during acceleration ($n^+ \in [1,2]$) and braking ($n^- \in [1,2]$).
As a result, the module implements a normalized friction ellipse representation 
(Figure \ref{img:GG}) at the front and rear axles, allowing the regulation of the vehicle operating point according to the desired balance between lateral and longitudinal performance.

Depending on temperature, the tire delivers an effective peak grip $D_i(T)$, 
which is generally lower than the nominal fully warmed value $D_i$. 
This effect can be expressed as a normalized performance coefficient:

\begin{small}
\begin{equation}\label{eq:K_T}
K_i(T) = \frac{D_i(T)}{D_{i}}
\end{equation}
\end{small}

\noindent

The function $D_i(T)$ can be identified offline from experimental data, 
allowing the construction of a temperature–performance map.
Accordingly, the Mission Planner can operate in two modes:

\begin{enumerate}
    \item \textbf{Model adaptation mode:} the performance target is kept at full potential ($K_i = 1$), while the tire model parameters are updated online exploiting the Grip Estimation module.
    \item \textbf{Target-scaling mode:} tire parameters remain unchanged, and the performance targets are directly scaled using the normalized gain $K_i(T)$ defined in Eq.\eqref{eq:K_T}. These scaled targets are subsequently applied to compute the admissible vehicle accelerations, as detailed in in Section \ref{sub_sec:l_plan}.
\end{enumerate}

From an operational standpoint, both approaches lead to an equivalent outcome: the longitudinal planner generates a reference speed profile consistent with the updated vehicle performance limits. Although the first approach is more physically intuitive, the second one is adopted during the competition.
In this configuration, vehicle parameters remain unchanged, and performance is regulated directly through the normalized gains $K_i(T)$, derived either from an updated parameter estimation or from a calibrated ideal progression.
This strategy avoids continuous modification of the tire model parameters within the prediction model. Since the planner and controller modules (Sec.~\ref{sub_sec:mppc}) rely on an optimization-based formulation, untested variations in model parameters may affect numerical conditioning and solver stability. Maintaining fixed model parameters while scaling performance targets ensures greater robustness and predictability within the optimization.

\subsection{Longitudinal Planner}\label{sub_sec:l_plan}

The longitudinal planner converts the normalized friction-ellipse performance targets provided by the Mission Performance into feasible velocity references.
The velocity profile along the path is computed using a forward–backward integration scheme using 
a point-mass formulation, where the admissible speed is constrained by path curvature and by the maximum longitudinal ($a^{\max}_{x}$) and lateral ($a^{\max}_{y}$) accelerations:
  \begin{align}
    \begin{cases}\label{eq:axBounds}
      a^{\max}_{x} = \frac{F_{x,\max} \cdot K_x}{m}, \\
      a^{\max}_{y} = \frac{F_{y,\max} \cdot K_y}{m}.
    \end{cases}
  \end{align}
Maximum longitudinal ($F_{x,\max}$) and lateral ($F_{y,\max}$) forces are derived from available tire grip, vehicle mass, and vertical tire loads according to the friction law:
\begin{small}
  \begin{align}
    \begin{cases}
      F_{x,\max} &= D_{xf} \, F_{zf} + D_{xr} \, F_{zr}, \\
      F_{y,\max} &= D_{yf} \, F_{zf} + D_{yr} \, F_{zr}.
    \end{cases} \label{eq:FxBounds}
  \end{align}
\end{small}

To describe Maximum combined-slip conditions, the admissible longitudinal and lateral 
accelerations are constrained through a generalized friction ellipse:
\begin{equation}\label{axComb}
  \left( \frac{a_x}{a_x^{\max}} \right)^n + 
  \left( \frac{a_y}{a_y^{\max}} \right)^n \le 1,
\end{equation}
where the exponent $n \in \{n_x, n_y\}$
is tuned by the Mission Performance module.
To further increase flexibility, separate exponents are introduced 
for the longitudinal and lateral contributions 
($n_x^+, n_y^+, n_x^-, n_y^-$), allowing independent shaping of the 
combined-slip behavior along each axis.

\subsection{Model Predictive Planning and Control (MPPC)}
\label{sub_sec:mppc}
In our framework, both the low-level planner and the controller are formulated as optimization problems. %\cite{raji_phd}. 
The Model Predictive Planner (MPP) optimizes the global path and the longitudinal speed profile generated by the longitudinal planner, accounting for deviations from the nominal trajectory and providing a high-fidelity reference for the Model Predictive Controller (MPC).
Both modules share the same vehicle model and parameters, ensuring consistency between planning and control.

	%%%%%%%%%%%%%%%%%%%%%%%%%%%%%%%%%%%%%%%%%%%%%%%%%%%%%%%%%%%%%%%%%%
	\section{Design and Implementation}
	\label{sec:designandimplementation}      
        The Mission Performance design is formed by the definition of specific track sectors, safety and stability metrics, and a progression updating logic.
        \subsection{Track Sectors}
        \label{track_sectors}

        \begin{figure}[b]
          \centering
          %\vspace{7pt} % Adds a small margin above the figure
          \includegraphics[width=0.7\columnwidth]{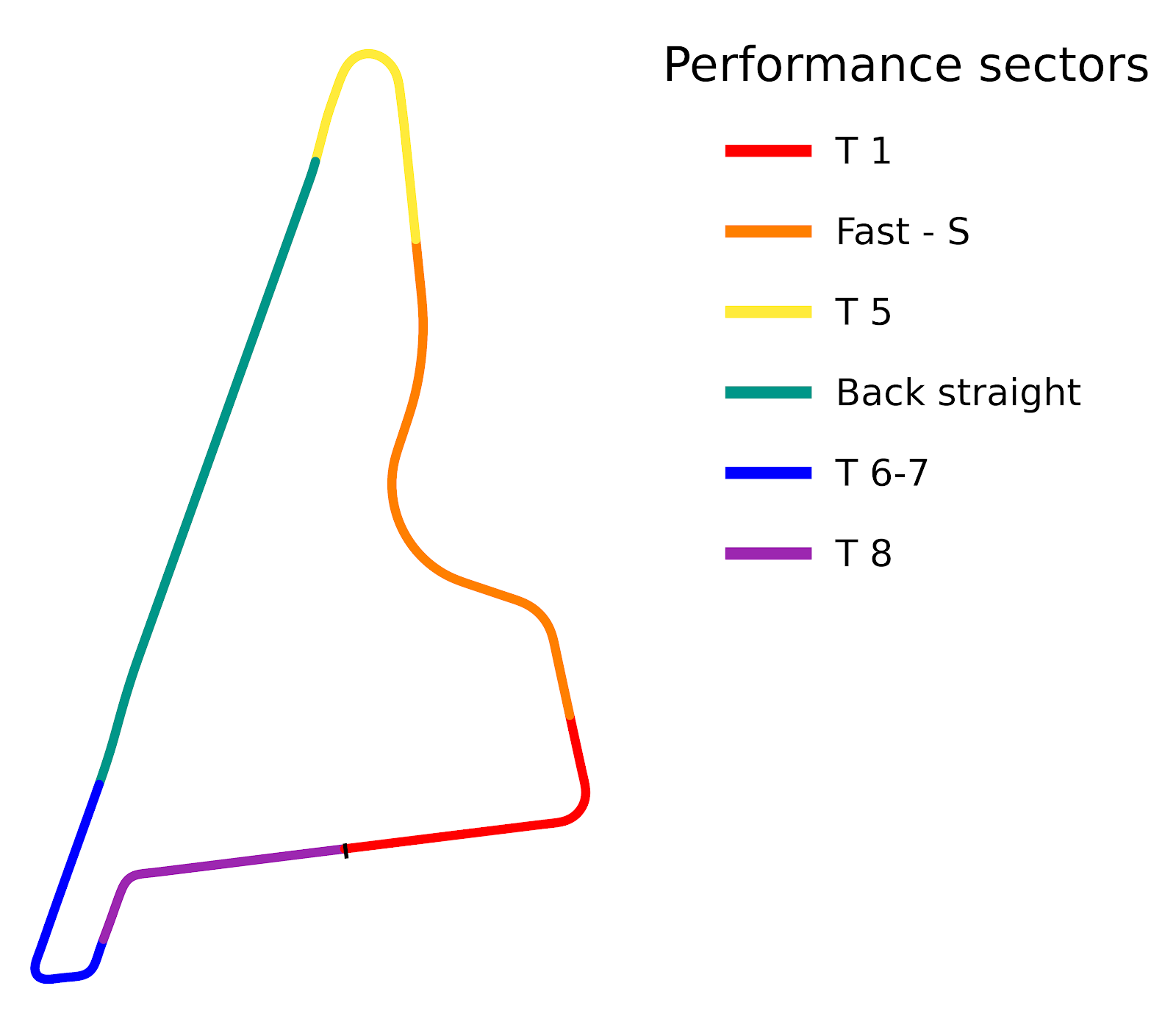}
          \caption{Performance sectors in Yas Marina circuit, north layout}
          \label{img:yas_north_performance_sectors_legend}
        \end{figure}
        
        The track is divided into sectors, each corresponding to a relevant turn or straight where a performance change is useful to minimize lap time while ensuring safety. Their ranges are described using the Frenet longitudinal coordinate (s) of a center reference line. Figure \ref{img:yas_north_performance_sectors_legend} shows the sectors used by the team during Qualification, Sprint Race, and Grand Finale events in the North Layout of the Yas Marina circuit. During these events, teams are not allowed to send any command to the car, which must operate fully autonomously from the moment it exits the pit lane until it returns to the pits at the end of the race. For this reason, a predefined increasing performance progression is assigned to each sector and performance type. In addition to the racing sectors, a dedicated pit lane zone is defined with static, low-performance limits. This specific profile is engaged when the vehicle is physically in the pit lane at the start of the event and during the in-lap when approaching the pit entry, for safety reasons related to the potential presence of humans.
        
        During race events, after the low-speed reconnaissance laps, the vehicle performances are gradually increased according to these progressions to warm up the tires and reach the best lap time as quickly as possible.
        The zones division is important to obtain the maximum performance safely. Track conditions are rarely uniform across the entire circuit. There can be variations in asphalt, track temperature (e.g. shaded areas), or the presence of debris that create heterogeneous grip levels. By managing performance on a per-sector basis, the system can adapt to local conditions without penalizing the entire lap. For instance, if a specific corner becomes slippery due to debris, the module detects the instability and reduces the performance only in that specific sector, while maintaining maximum pace in the unaffected sections. Conversely, a global strategy would force the vehicle to limit its performance based on the worst-case scenario of the most critical corner, resulting in a suboptimal lap time.

        \subsection{Logic and Implementation}
        The logic is divided into two steps:
        \begin{enumerate}
            \item Metrics computation: this step is executed at a frequency of 20 Hz. It computes the safety and vehicle dynamics metrics that are used to tune the performances. These metrics are:
            
            \begin{itemize}
                \item lateral error: the path tracking error of the controller
                \item understeer angle \cite{guiggiani}: representing an oversteering behaviour when negative and understeering when positive. For simplicity, for the remainder of this paper, they will be named Oversteer angle and Understeer angle.
                \item lateral velocity
                \item longitudinal acceleration and slip
                \item Traction Control (TC) / Anti-lock Braking System (ABS) activations
            \end{itemize}

            A moving average filter is applied to the lateral error, understeer degree, and lateral velocity estimations to improve robustness against measurement noise. The filtered value $\bar{x}$ is computed over a window of $N=10$ samples at a frequency $f_s = 20$ Hz, resulting in a time window of $T_w = 0.5$ s. This prevents transient disturbances from triggering a decrease of performance.
            
            \item Performance update: this step is executed each time the vehicle enters a new sector to update the performance parameters of the just completed sector for the subsequent lap. This update is skipped under the following conditions:
            \begin{itemize}
                \item the car is in the pit lane
                \item the car is completing warm-up laps
                \item there is a race control flag that is limiting the speed of the car (e.g \textit{YELLOW FLAG} following an incident)
            \end{itemize}
        \end{enumerate}
        
        The Mission Performance module automatically manages the following performance parameters:
        \begin{itemize} 
            \item $K_{y}$, $K_{x+}$ and $K_{x-}$: They dictate how much lateral, longitudinal acceleration and braking performance force can be sustained;
            \item $a^{\max}_{x}$ and $a^{\min}_{x}$: respectively, maximum target acceleration and deceleration;
            \item $n^+$($n^+_x$,$n^+_y$) and $n^-$($n^-_x$,$n^-_x$): They dictate how much longitudinal force can be sustained while cornering, effectively regulating both corner-exit traction (\textit{$n^+$}) and trail-braking aggressiveness (\textit{$n^-$}).
        \end{itemize}
        
        For each performance parameter, there is a predefined step-wise progression. Based on the computed safety and vehicle dynamics metrics, each performance parameter can be:
        \begin{itemize}
            \item \textbf{Reduced}: if the metric value exceeds its predefined threshold, the associated performance parameters are decreased. Instead of fully reverting to the previous discrete level, the system reduces the performance by half of the previous increment. This step-halving approach prevents oscillations between two consecutive levels and efficiently converges toward the highest safe performance limit. However, to prevent the decrement from becoming infinitesimally small, a tunable minimum reduction step is defined for each parameter. This ensures that, after a certain number of halving iterations, the performance will eventually step down to the previous discrete level in the progression.
            \item \textbf{Maintained}: if the metric value is between $90\%$ and $100\%$ of the threshold, the current performance level is maintained. However, if the absolute value of the performance parameter exceeds a conservative upper bound, it is reduced anyway to prevent the vehicle from operating continuously at the edge of instability.
            \item \textbf{Increased}: if the metric value remains below $90\%$ of the threshold, the performance parameters are increased by one progression step.
        \end{itemize}

        Table \ref{tab:metrics_performance} lists the metrics, the affected performance parameters, and their corresponding thresholds. Some thresholds are tuned slightly differently for specific turns.        
        \begin{table}[h]
            \centering
            \caption{Safety and Vehicle Dynamics Performance Metrics}
            \label{tab:metrics_performance}
            % Il comando resizebox adatta il contenuto alla larghezza della colonna (\columnwidth)
            \resizebox{\columnwidth}{!}{%
                \begin{tabular}{l c l}
                    \toprule
                    \textbf{Metric} & \textbf{Threshold} & \textbf{Affected Performance Type} \\
                    \midrule
                    Lateral error & $1.0\,\mathrm{m}$ & $K_y$, $n^+$, $n^-$ \\
                    Understeer angle & $0.06\,\mathrm{rad}$ & $K_y$, $n^+$, $n^-$ \\
                    Oversteer angle & $0.04\,\mathrm{rad}$ & $K_y$, $n^+$, $n^-$ \\
                    Lateral velocity & $3.0\,\mathrm{m/s}$ & $K_y$, $n^+$, $n^-$ \\
                    Longitudinal acceleration slip & $20\,\%$ & $K_{x+}$, $a^{\max}_{x}$, $n^+$ \\
                    TC activations & $100$ & $K_{x+}$, $a^{\max}_{x}$, $n^+$ \\
                    Longitudinal brake slip & $50\,\%$ & $K_{x-}$, $a^{\min}_{x}$, $n^-$ \\
                    ABS activations & $100$ & $K_{x-}$, $a^{\min}_{x}$, $n^-$ \\
                    \bottomrule
                \end{tabular}%
            }
        \end{table}

        To ensure vehicle stability, the following constraints enforce a safety hierarchy: a solid baseline of pure lateral grip must be established before demanding high combined tire forces, and conversely, combined aggressiveness is always reduced first if instability occurs:
        \begin{itemize}
            \item $n^+$ and $n^-$ can only be increased if $K_y$ is already greater than a threshold (tuned differently for each sector)
            \item $K_y$ can only be decreased if $n^+$ and $n^-$ are at their minimum
            \item $K_{x+}$ and $K_{x-}$ can only be decreased if $n^+$ and $n^-$ are at their minimum
            
        \end{itemize}

        In addition, two constraints are used to clamp the maximum performance:
        \begin{itemize}
            \item Static maximum clamp: for every performance parameter, there is a maximum value that can be reached, which is tuned based on offline test data.
            \item Tire temperature clamp: a LUT maps the front tire’s internal temperature to maximum safe lateral and longitudinal performance limits, derived from experimental data. This mechanism ensures safety during sub-optimal thermal conditions, such as when tires cool down due to reduced speeds. A practical example is a yellow flag scenario, where the vehicle is speed-limited. By enforcing these temperature-dependent limits, the system ensures that the performance progression re-initializes from a safe baseline when racing resumes.
        \end{itemize}

        Algorithm \ref{alg:perf_update} outlines the dynamic performance update loop. Getting the needed information from the state estimations, the Mission Performance evaluates whether any metric exceeds its corresponding safety threshold (or reaches at least 90\% of that limit) within the current sector. A safety clamp based on tire temperatures is then applied. Upon entering a new track zone, provided the vehicle is in an active racing state, the counters from the recently completed sector are used to update the global counters for consecutive exceedance. The performance parameters are subsequently adjusted based on these updated records. Finally, a static maximum clamp is enforced, and the local zone counters are reset in preparation for the next sector. 

        \begin{algorithm}[htbp]
        \small % <-- Font del codice e dei commenti
        \caption{Dynamic performance update loop}
        \label{alg:perf_update}
        \SetAlgoLined
        
        % Stile commenti: solo corsivo, manterranno la stessa grandezza del codice (\small)
        \newcommand\mycommfont[1]{\itshape #1}
        \SetCommentSty{mycommfont}
        
        \KwIn{Vehicle State, Race Control flags}
        \KwOut{Updated vehicle performance scaling parameters}
        \BlankLine
        \SetKwFunction{FMain}{UpdatePerf}
        \SetKwProg{Fn}{Procedure}{:}{}
        \Fn{\FMain{}}{
            RetrieveSafetyAndVehicleDynamicsMetrics()\;
            \BlankLine
            \ForEach{metric}{
                \ForEach{performance parameter related to metric}{
                    UpdateCurrentSectorExceedanceCounters()\;
                }
            }
            clamPerfParam(current\_sector, tire\_temp)
            \BlankLine
            is\_active\_racing $\leftarrow$ (\textbf{not} in\_pit) \textbf{and} (\textbf{not} warmup) \textbf{and} (\textbf{not} race\_control\_flags)\;
            \BlankLine
            \If{entered\_new\_sector \textbf{and} is\_active\_racing}{
                \ForEach{performance parameter}{
                    UpdateConsecutiveLapsCounters(previous\_sector)\;
                    UpdatePerfParam(previous\_sector)\;
                    ClampPerfParam(previous\_sector, static\_max\_limit)\;
                }
                ResetThresholdExceedanceCounters()\;
            }
        }
        \end{algorithm}

        During testing sessions, a mode is available to manually override performance limits for specific sectors. This functionality facilitates the targeted testing of algorithmic updates within specific track segments.

        \definecolor{pgreen}{HTML}{009900}

        \begin{table*}[htbp!]
            \centering
            \caption{Lap-by-Lap Performance Progression per Sector}
            \label{tab:lap_progression}
            \resizebox{\textwidth}{!}{
            \begin{tabular}{@{} l l *{13}{c} @{}}
            \toprule
            \textbf{Zone} & \textbf{Parameter} & \textbf{Lap -1} & \textbf{Lap 0} & \textbf{Lap 1} & \textbf{Lap 2} & \textbf{Lap 3} & \textbf{Lap 4} & \textbf{Lap 5} & \textbf{Lap 6} & \textbf{Lap 7} & \textbf{Lap 8} & \textbf{Lap 9} & \textbf{Lap 10} & \textbf{Lap 11} \\ \midrule
            
            % ZONE T1
            \multirow{5}{*}{T1} 
            & $K_y$ & 0.9 & 0.9 & 0.9 & 0.95 & 0.97 & \textcolor{pgreen}{1.0} & \textcolor{pgreen}{1.0} & \textcolor{pgreen}{1.0} & \textcolor{pgreen}{1.0} & \textcolor{pgreen}{1.0} & \textcolor{pgreen}{1.0} & \textcolor{pgreen}{1.0} & \textcolor{pgreen}{1.0} \\
            & $n^+_x$ & 1.2 & 1.2 & 1.2 & 1.2 & 1.2 & 1.4 & 1.6 & 1.8 & \textcolor{pgreen}{2.0} & \textcolor{pgreen}{2.0} & \textcolor{pgreen}{2.0} & \textcolor{pgreen}{2.0} & \textcolor{pgreen}{2.0} \\
            & $n^+_y$ & 1.0 & 1.0 & 1.0 & 1.0 & 1.0 & 1.05 & 1.1 & \textcolor{pgreen}{1.2} & \textcolor{pgreen}{1.2} & \textcolor{pgreen}{1.2} & \textcolor{pgreen}{1.2} & \textcolor{pgreen}{1.2} & \textcolor{pgreen}{1.2} \\
            & $n^-_x$ & 1.0 & 1.0 & 1.0 & 1.0 & 1.0 & 1.2 & 1.4 & 1.6 & \textcolor{pgreen}{1.8} & \textcolor{pgreen}{1.8} & \textcolor{pgreen}{1.8} & \textcolor{pgreen}{1.8} & \textcolor{pgreen}{1.8} \\
            & $n^-_y$ & 1.0 & 1.0 & 1.0 & 1.0 & 1.0 & 1.05 & \textcolor{pgreen}{1.1} & \textcolor{pgreen}{1.1} & \textcolor{pgreen}{1.1} & \textcolor{pgreen}{1.1} & \textcolor{pgreen}{1.1} & \textcolor{pgreen}{1.1} & \textcolor{pgreen}{1.1} \\ \midrule
            
            % ZONE Fast S
            \multirow{5}{*}{Fast S} 
            & $K_y$ & 0.9 & 0.9 & 0.9 & 0.73 & 0.75 & \textcolor{pgreen}{1.0} & \textcolor{pgreen}{1.0} & \textcolor{pgreen}{1.0} & \textcolor{pgreen}{1.0} & \textcolor{pgreen}{1.0} & \textcolor{pgreen}{1.0} & \textcolor{pgreen}{1.0} & \textcolor{pgreen}{1.0} \\
            & $n^+_x$ & \textcolor{pgreen}{1.2} & \textcolor{pgreen}{1.2} & \textcolor{pgreen}{1.2} & \textcolor{pgreen}{1.2} & \textcolor{pgreen}{1.2} & \textcolor{pgreen}{1.2} & \textcolor{pgreen}{1.2} & \textcolor{pgreen}{1.2} & \textcolor{pgreen}{1.2} & \textcolor{pgreen}{1.2} & \textcolor{pgreen}{1.2} & \textcolor{pgreen}{1.2} & \textcolor{pgreen}{1.2} \\
            & $n^+_y$ & \textcolor{pgreen}{1.0} & \textcolor{pgreen}{1.0} & \textcolor{pgreen}{1.0} & \textcolor{pgreen}{1.0} & \textcolor{pgreen}{1.0} & \textcolor{pgreen}{1.0} & \textcolor{pgreen}{1.0} & \textcolor{pgreen}{1.0} & \textcolor{pgreen}{1.0} & \textcolor{pgreen}{1.0} & \textcolor{pgreen}{1.0} & \textcolor{pgreen}{1.0} & \textcolor{pgreen}{1.0} \\
            & $n^-_x$ & 1.0 & 1.0 & 1.0 & 1.0 & 1.0 & 1.0 & 1.0 & \textcolor{pgreen}{1.2} & \textcolor{pgreen}{1.2} & \textcolor{pgreen}{1.2} & \textcolor{pgreen}{1.2} & \textcolor{red}{1.2} & \textcolor{red}{1.1} \\
            & $n^-_y$ & 1.0 & 1.0 & 1.0 & 1.0 & 1.0 & 1.0 & 1.05 & 1.1 & \textcolor{pgreen}{1.2} & \textcolor{pgreen}{1.2} & \textcolor{pgreen}{1.2} & \textcolor{red}{1.2} & \textcolor{red}{1.1} \\ \midrule
            
            % ZONE T5
            \multirow{5}{*}{T5} 
            & $K_y$ & 0.9 & 0.9 & 0.9 & 0.95 & 0.97 & \textcolor{pgreen}{1.0} & \textcolor{pgreen}{1.0} & \textcolor{pgreen}{1.0} & \textcolor{pgreen}{1.0} & \textcolor{pgreen}{1.0} & \textcolor{pgreen}{1.0} & \textcolor{pgreen}{1.0} & \textcolor{pgreen}{1.0} \\
            & $n^+_x$ & 1.2 & 1.2 & 1.2 & 1.4 & 1.6 & 1.8 & \textcolor{pgreen}{2.0} & \textcolor{pgreen}{2.0} & \textcolor{pgreen}{2.0} & \textcolor{pgreen}{2.0} & \textcolor{pgreen}{2.0} & \textcolor{pgreen}{2.0} & \textcolor{pgreen}{2.0} \\
            & $n^+_y$ & 1.0 & 1.0 & 1.0 & 1.05 & 1.1 & 1.2 & 1.4 & 1.6 & 1.8 & \textcolor{pgreen}{2.0} & \textcolor{pgreen}{2.0} & \textcolor{pgreen}{2.0} & \textcolor{pgreen}{2.0} \\
            & $n^-_x$ & 1.0 & 1.0 & 1.0 & 1.2 & 1.4 & 1.6 & 1.8 & \textcolor{pgreen}{2.0} & \textcolor{pgreen}{2.0} & \textcolor{pgreen}{2.0} & \textcolor{pgreen}{2.0} & \textcolor{pgreen}{2.0} & \textcolor{pgreen}{2.0} \\
            & $n^-_y$ & 1.0 & 1.0 & 1.0 & 1.05 & 1.1 & 1.2 & 1.4 & \textcolor{pgreen}{1.5} & \textcolor{pgreen}{1.5} & \textcolor{pgreen}{1.5} & \textcolor{pgreen}{1.5} & \textcolor{pgreen}{1.5} & \textcolor{pgreen}{1.5} \\ \midrule
            
            % ZONE Back Straight
            %\multirow{5}{*}{\shortstack[l]{Back \\ Straight}}
            %& $K_y$ & 0.9 & 0.9 & 0.9 & 0.95 & 0.97 & \textcolor{pgreen}{1.0} & \textcolor{pgreen}{1.0} & \textcolor{pgreen}{1.0} & \textcolor{pgreen}{1.0} & \textcolor{pgreen}{1.0} & \textcolor{pgreen}{1.0} & \textcolor{pgreen}{1.0} & \textcolor{pgreen}{1.0} \\
            %& $n^+_x$ & 1.2 & 1.2 & 1.2 & 1.4 & 1.6 & 1.8 & \textcolor{pgreen}{2.0} & \textcolor{pgreen}{2.0} & \textcolor{pgreen}{2.0} & \textcolor{pgreen}{2.0} & \textcolor{pgreen}{2.0} & \textcolor{pgreen}{2.0} & \textcolor{pgreen}{2.0} \\
            %& $n^+_y$ & 1.0 & 1.0 & 1.0 & 1.1 & 1.1 & 1.2 & 1.4 & 1.6 & 1.8 & \textcolor{pgreen}{2.0} & \textcolor{pgreen}{2.0} & \textcolor{pgreen}{2.0} & \textcolor{pgreen}{2.0} \\
            %& $n^-_x$ & 1.0 & 1.0 & 1.0 & 1.2 & 1.4 & 1.6 & 1.8 & \textcolor{pgreen}{2.0} & \textcolor{pgreen}{2.0} & \textcolor{pgreen}{2.0} & \textcolor{pgreen}{2.0} & \textcolor{pgreen}{2.0} & \textcolor{pgreen}{2.0} \\
            %& $n^-_y$ & 1.0 & 1.0 & 1.0 & 1.1 & 1.1 & 1.2 & 1.4 & 1.6 & 1.8 & \textcolor{pgreen}{2.0} & \textcolor{pgreen}{2.0} & \textcolor{pgreen}{2.0} & \textcolor{pgreen}{2.0} \\ \midrule
            
            % ZONE T6-T7
            \multirow{5}{*}{T6-T7} 
            & $K_y$ & 0.9 & 0.9 & 0.9 & 0.95 & 0.97 & \textcolor{pgreen}{1.0} & \textcolor{pgreen}{1.0} & \textcolor{pgreen}{1.0} & \textcolor{pgreen}{1.0} & \textcolor{pgreen}{1.0} & \textcolor{pgreen}{1.0} & \textcolor{pgreen}{1.0} & \textcolor{pgreen}{1.0} \\
            & $n^+_x$ & 1.2 & 1.2 & 1.2 & 1.4 & 1.6 & 1.8 & \textcolor{pgreen}{2.0} & \textcolor{pgreen}{2.0} & \textcolor{red}{2.0} & \textcolor{red}{1.9} & \textcolor{red}{1.85} & \textcolor{red}{1.83} & \textcolor{red}{1.81} \\
            & $n^+_y$ & 1.0 & 1.0 & 1.0 & 1.05 & 1.1 & 1.2 & 1.4 & \textcolor{pgreen}{1.6} & \textcolor{red}{1.6} & \textcolor{red}{1.5} & \textcolor{red}{1.45} & \textcolor{red}{1.43} & \textcolor{red}{1.41} \\
            & $n^-_x$ & 1.0 & 1.0 & 1.0 & 1.2 & 1.4 & 1.6 & 1.8 & \textcolor{pgreen}{2.0} & \textcolor{red}{2.0} & \textcolor{red}{1.9} & \textcolor{red}{1.85} & \textcolor{red}{1.83} & \textcolor{red}{1.81} \\
            & $n^-_y$ & 1.0 & 1.0 & 1.0 & 1.05 & 1.1 & 1.2 & 1.4 & \textcolor{pgreen}{1.6} & \textcolor{red}{1.6} & \textcolor{red}{1.5} & \textcolor{red}{1.45} & \textcolor{red}{1.43} & \textcolor{red}{1.41} \\ \midrule
            
            % ZONE T8
            \multirow{5}{*}{T8} 
            & $K_y$ & 0.9 & 0.9 & 0.95 & 0.97 & \textcolor{pgreen}{1.0} & \textcolor{pgreen}{1.0} & \textcolor{pgreen}{1.0} & \textcolor{pgreen}{1.0} & \textcolor{pgreen}{1.0} & \textcolor{pgreen}{1.0} & \textcolor{pgreen}{1.0} & \textcolor{pgreen}{1.0} & \textcolor{pgreen}{1.0} \\ 
            & $n^+_x$ & 1.2 & 1.2 & 1.2 & 1.2 & 1.4 & 1.6 & 1.8 & \textcolor{pgreen}{2.0} & \textcolor{pgreen}{2.0} & \textcolor{pgreen}{2.0} & \textcolor{pgreen}{2.0} & \textcolor{pgreen}{2.0} & \textcolor{pgreen}{2.0} \\
            & $n^+_y$ & 1.0 & 1.0 & 1.0 & 1.0 & 1.05 & 1.1 & \textcolor{pgreen}{1.2} & \textcolor{pgreen}{1.2} & \textcolor{pgreen}{1.2} & \textcolor{pgreen}{1.2} & \textcolor{pgreen}{1.2} & \textcolor{pgreen}{1.2} & \textcolor{pgreen}{1.2} \\
            & $n^-_x$ & 1.0 & 1.0 & 1.0 & 1.0 & 1.2 & 1.4 & 1.6 & 1.8 & \textcolor{pgreen}{2.0} & \textcolor{pgreen}{2.0} & \textcolor{pgreen}{2.0} & \textcolor{pgreen}{2.0} & \textcolor{pgreen}{2.0} \\
            & $n^-_y$ & 1.0 & 1.0 & 1.0 & 1.0 & 1.05 & 1.1 & 1.2 & 1.4 & 1.6 & 1.8 & \textcolor{pgreen}{2.0} & \textcolor{pgreen}{2.0} & \textcolor{pgreen}{2.0} \\ \midrule
            
            % FINAL AVERAGES AND LAP TIME
            \multicolumn{2}{@{}l}{\textbf{Avg Temp Front [$^\circ$C]}} & 60.5 & 55.7 & 54.0 & 56.2 & 61.5 & 67.3 & 72.2 & 75.8 & 79.0 & 81.2 & 82.7 & 83.7 & 84.0 \\
            \multicolumn{2}{@{}l}{\textbf{Avg Temp Rear [$^\circ$C]}}  & 68.7 & 58.8 & 51.8 & 49.3 & 50.3 & 51.8 & 53.3 & 54.8 & 56.0 & 56.5 & 57.8 & 58.7 & 59.8 \\
            \multicolumn{2}{@{}l}{\textbf{Lap Time [s]}}               & 102.39 & 105.71 & 61.58 & 60.30 & 59.68 & 58.97 & 58.78 & 61.08 & 59.07 & 58.94 & 59.17 & 59.31 & 59.09 \\ \bottomrule
            
            \multicolumn{15}{p{\textwidth}}{\footnotesize \textit{Note:} Green values highlight that the performance parameter reached the maximum clamp in that sector. Red values highlight that a performance decrease was triggered for the following lap. The parameters that are not reported were set to maximum from the beginning and remained constant.} \\
            \end{tabular}
            }
        \end{table*}

        \begin{figure*}[htbp]
            \centering
            \includegraphics[width=\linewidth]{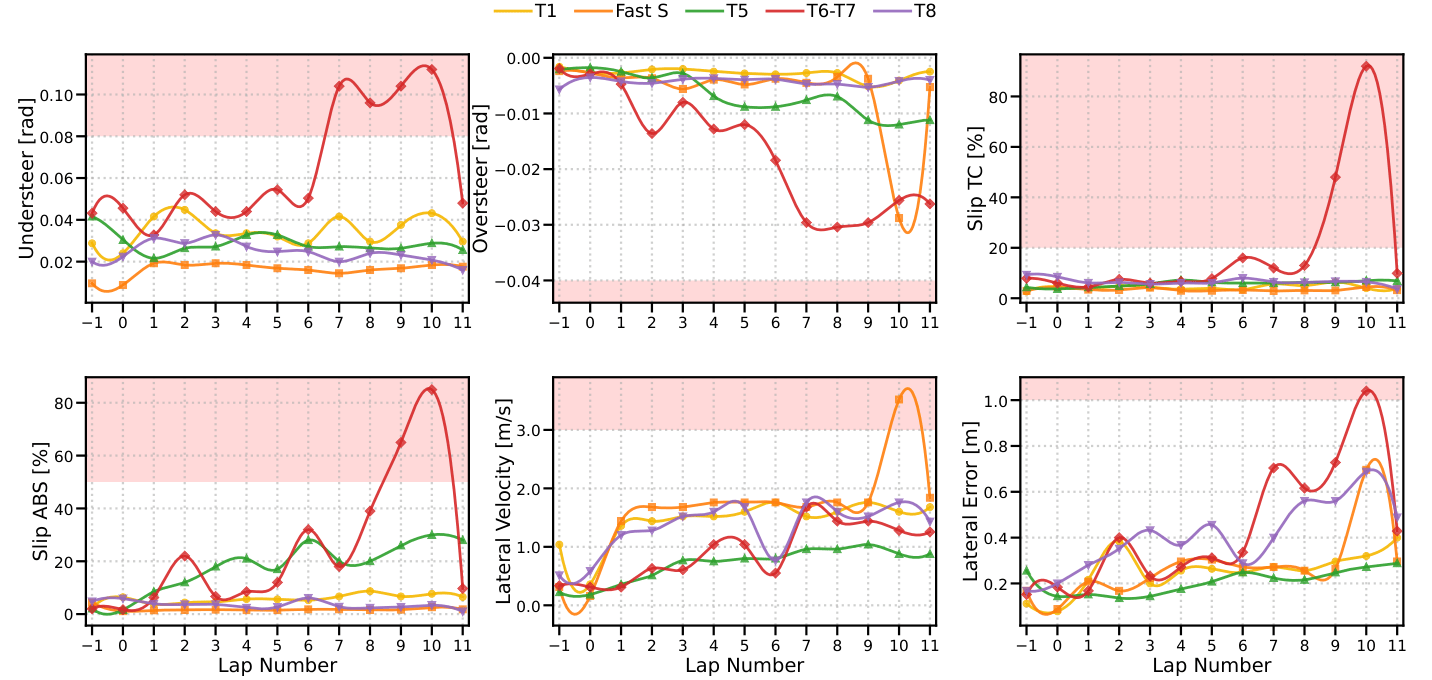}
            \caption{Lap-by-lap evolution of the main safety metrics in the different sectors. Red shaded regions indicate values that exceed their respective safety thresholds.}
            \label{img:metrics_evolution}
        \end{figure*}
        
        %%%%%%%%%%%%%%%%%%%%%%%%%%%%%%%%%%%%%%%%%%%%%%%%%%%%%%%%%%%%%%%%%%
        \section{Results}
	\label{sec:results}
        
        The Mission Performance module was deployed during the A2RL Season 2 competition at the Yas Marina Circuit in Abu Dhabi in October and November 2025, covering the Qualifying, Sprint Race, and Grand Finale sessions. The module is highly computationally efficient. By delegating the resource-intensive grip and state estimation tasks to dedicated upstream modules, the core decision-making logic executes in under 0.2 ms on an Intel Core i9-14900HX processor.

        \subsection{Race Progression}
	\label{sec:results_race}        
        Table  \ref{tab:lap_progression} and Figure \ref{img:metrics_evolution} illustrate the system's behavior during the first 11 laps of the Grand Finale, held on November 15th, 2025. The event featured a six-car grid with a rolling start procedure. It can be noticed that during the first two reconnaissance laps the performance parameters have not been updated. The car's speed is initially limited, causing the tires, which were heated with tire warmers, to cool down slightly. After the first full-speed lap, the performance is gradually increased. The car received the  \textit{GREEN} flag signal before the start/finish line of the last reconnaissance lap, so it started to increase the performances in \textit{T8} of lap 1 (the first one at full speed). The lateral performance progression is uniform across all sectors, whereas the combined performance progression depends on the minimum lateral performance threshold required to initiate increases in the combined parameters, that varies by sector.

        The sixth lap of the race was more than 2 seconds slower than the previous one due to a lapping maneuver that resulted in a loss of time. As a consequence, the trailing vehicle reduced the gap to under 15 meters, mandating a yield of the racing line as per competition regulations. Specifically, the vehicle was required to concede at least 3.5 meters of clearance from the track boundary on the overtaking side. As depicted in Figure \ref{img:metrics_evolution}, the required evasive maneuver triggered the understeer safety threshold, resulting in a reduction of the combined performance parameters. However, this intervention was caused by a non-nominal situation where the car was forced to deviate from the ideal racing trajectory. The subsequent loss of pace allowed the trailing opponent to remain within the 15-meter proximity window throughout that sector, forcing our vehicle to continuously yield the right of way for the following four laps.
        
        Another performance reduction occurred in the \textit{Fast S} sector during lap 10, triggered by an erroneous Inertial Measurement Unit (IMU) reading that lasted for approximately half a second. This anomaly resulted in inaccurate estimations of both longitudinal and lateral velocities. Under these circumstances, the automated performance reduction functioned exactly as intended, serving as a critical safety measure. This problem demonstrated that unforeseen anomalies in vehicle state estimation and track conditions can be safely managed autonomously during a race, without requiring human intervention.

    \subsection{Oversteering Scenario}
    \label{sec:results_oversteer}        
\begin{figure}[t]
          \centering
          %\vspace{7pt} % Adds a small margin above the figure
          \includegraphics[width=1.0\columnwidth]{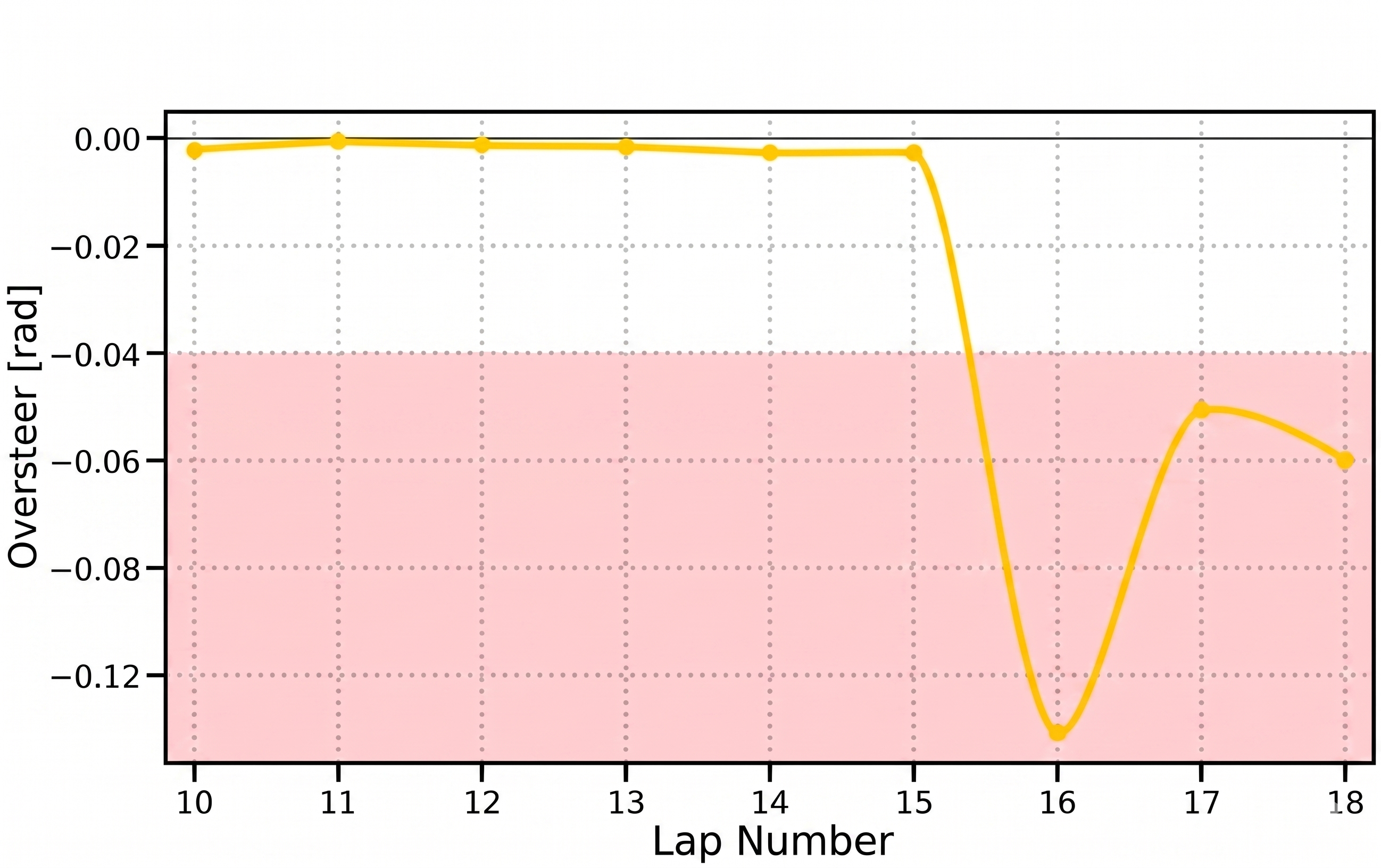}
          \caption{Oversteer in T1 during a test day}
          \label{img:t1_oversteer}
    \end{figure}

    Adaptability to oversteering scenarios is demonstrated during a track test day, where this condition was encountered in sector T1 during the final three laps running on an old set of tires. Table \ref{tab:t1_progression} details the last part of the progression of that sector, where performance parameters were decreased in response to the oversteer, as illustrated in Figure \ref{img:t1_oversteer}. The performance targets followed the nominal progression until lap 8 and maintained the maximum clamp values until lap 14. Subsequently, a reduction was triggered by the oversteering metrics during Laps 15, 16, and 17, before the conclusion of the run and the pit-in request by Race Control. Thanks to this reaction, the value has been drastically reduced, bringing it close to the threshold. Nevertheless, this scenario highlights the high dependency on the tuning of the reduction step and the thresholds, which in this case were not conservative enough and kept the vehicle still in a not fully safe stability.

    \begin{table}[t]
        \centering
        \caption{Progression in T1 during a test day}
        \label{tab:t1_progression}
        \small % Riduce leggermente il font per garantire che stia in una colonna
        \begin{tabular}{@{} l l c c c c @{}}
        \toprule
        \textbf{Zone} & \textbf{Parameter} & \textbf{Lap 8-14} & \textbf{Lap 15} & \textbf{Lap 16} & \textbf{Lap 17} \\ \midrule
        
        % ZONE T1
        \multirow{5}{*}{T1} 
        & $K_y$ & \textcolor{pgreen}{1.0} & \textcolor{pgreen}{1.0} & \textcolor{pgreen}{1.0} & \textcolor{pgreen}{1.0} \\
        & $n^+_x$   & \textcolor{pgreen}{2.0} & \textcolor{red}{2.0}    & \textcolor{red}{1.9}    & \textcolor{red}{1.85} \\
        & $n^+_y$   & \textcolor{pgreen}{1.2} & \textcolor{red}{1.2}    & \textcolor{red}{1.15}   & \textcolor{red}{1.125} \\
        & $n^-_x$   & \textcolor{pgreen}{2.0} & \textcolor{red}{2.0}    & \textcolor{red}{1.9}    & \textcolor{red}{1.85} \\
        & $n^-_y$   & \textcolor{pgreen}{1.2} & \textcolor{red}{1.2}    & \textcolor{red}{1.15}   & \textcolor{red}{1.125} \\ \bottomrule
        
        \end{tabular}
    \end{table}

    %%%%%%%%%%%%%%%%%%%%%%%%%%%%%%%%%%%%%%%%%%%%%%%%%%%%%%%%%%%%%%%%%%
    \section{Conclusion and Future Work}
    \label{sec:conclusion}
        In this paper, we presented the Mission Performance module developed for our autonomous racing stack. The system proved to be a robust solution for managing the delicate trade-off between vehicle performance and safety. By dynamically adapting longitudinal, lateral, and combined performance parameters on a per-sector basis, the module efficiently automated the initial warm-up phase and reacted in real-time to track and vehicle conditions without human intervention.

        Future work will focus on refining the system's response to multi-agent scenarios. Specifically, a tighter integration between the planning and performance modules is necessary to better manage yields and overtaking situations. By having the planner explicitly signal an evasive or overtaking maneuver, the Mission Performance module should be able to temporarily discard data that reflects isolated cases rather than clear dynamic limits, thereby preventing unwarranted performance reductions.

        Furthermore, we aim to completely replace the static progression vectors by heavily exploiting real-time tire temperatures. Currently, the temperature LUT acts as a safety clamp. In the future, it will evolve into an active exploratory progression mechanism. This dynamic LUT will be designed with boundaries that are inherently safe, yet permissive enough to allow the vehicle to push its limits, generate tire heat, and naturally advance to higher performance tiers. This development aligns with the ultimate goal of the Mission Performance module: minimizing manual tuning and automating the entire performance management process. By relying on a temperature-driven exploratory LUT, the only required manual input would be the selection of high-level aggressiveness profiles tailored to the specific vehicle conditions, such as new versus used tire compounds.
        Finally, a close interaction with the Grip Estimation module will be explored to exploit its information and potentially build a hybrid mode in which the model parameters can be updated if the actual mismatch is higher than a certain threshold.
    
    %%%%%%%%%%%%%%%%%%%%%%%%%%%%%%%%%%%%%%%%%%%%%%%%%%%%%%%%%%%%%%%%%%
    \section*{ACKNOWLEDGMENTS}
    We would like to thank the entire Unimore Racing team for their support, collaboration, and valuable contributions throughout this work.

    %%%%%%%%%%%%%%%%%%%%%%%%%%%%%%%%%%%%%%%%%%%%%%%%%%%%%%%%%%%%%%%%%%
    %\addtolength{\textheight}{-12cm}
    %\vspace{10mm}
    \bibliographystyle{IEEEtran}
    % Your .bib file here
    \bibliography{root} 

@article{betz_survey,
  author    = {Betz, J. and Zheng, H. and Liniger, A. and Rosolia, U. and Karle, P. and Behl, M. and Krovi, V. and Mangharam, R.},
  title     = {Autonomous Vehicles on the Edge: A Survey on Autonomous Vehicle Racing},
  journal   = {IEEE Open Journal of Intelligent Transportation Systems},
  volume    = {3},
  pages     = {458--488},
  year      = {2022},
  doi       = {10.1109/OJITS.2022.3181510}
}

@misc{iac,
  author       = {Mitchell, P.},
  title        = {Indy Autonomous Challenge},
  howpublished = {\url{https://www.indyautonomouschallenge.com/}},
  year         = {2026}
}

@misc{a2rl,
  author       = {{ASPIRE UAE}},
  title        = {Abu Dhabi Autonomous Racing League},
  howpublished = {\url{https://a2rl.io/}},
  year         = {2026}
}

@article{fuel_minimization,
    title = {Optimal low-level control strategies for a high-performance hybrid electric power unit},
    journal = {Applied Energy},
    volume = {276},
    pages = {115248},
    year = {2020},
    issn = {0306-2619},
    doi = {https://doi.org/10.1016/j.apenergy.2020.115248},
    author = {Camillo Balerna and Nicolas Lanzetti and Mauro Salazar and Alberto Cerofolini and Christopher Onder}
}

@article{endurance_thermal_constraints,
    author = {van Kampen, Jorn and Herrmann, Thomas and Hofman, Theo and Salazar, Mauro},
    year = {2023},
    month = {01},
    pages = {1},
    title = {Optimal Endurance Race Strategies for a Fully Electric Race Car Under Thermal Constraints},
    volume = {PP},
    journal = {IEEE Transactions on Control Systems Technology},
    doi = {10.1109/TCST.2023.3340793}
}

@article{lap_time_minimization_electric,
  title={A Convex Optimization Framework for Minimum Lap Time Design and Control of Electric Race Cars},
  author={Borsboom, O. and Fahdzyana, C. A. and Hofman, T. and Salazar, M.},
  journal={IEEE Transactions on Vehicular Technology},
  volume={70},
  number={10},
  pages={9920--9931},
  year={2021},
  publisher={IEEE},
  doi={10.1109/TVT.2021.3106364}
}

@ARTICLE{minimum_lap_time_control,
  author={Salazar, Mauro and Duhr, Pol and Balerna, Camillo and Arzilli, Luca and Onder, Christopher H.},
  journal={IEEE Transactions on Vehicular Technology}, 
  title={Minimum Lap Time Control of Hybrid Electric Race Cars in Qualifying Scenarios}, 
  year={2019},
  volume={68},
  number={8},
  pages={7296-7308},
  doi={10.1109/TVT.2019.2920777}}

@ARTICLE{time_optimal_control,
  author={Salazar, Mauro and Elbert, Philipp and Ebbesen, Soren and Bussi, Carlo and Onder, Christopher H.},
  journal={IEEE Transactions on Control Systems Technology}, 
  title={Time-optimal Control Policy for a Hybrid Electric Race Car}, 
  year={2017},
  volume={25},
  number={6},
  pages={1921-1934},
  doi={10.1109/TCST.2016.2642830}}

@INPROCEEDINGS{equivalent_lap_time_minimization,
  author={Salazar, Mauro and Balerna, Camillo and Chisari, Eugenio and Bussi, Carlo and Onder, Christopher H.},
  booktitle={2018 IEEE Conference on Decision and Control (CDC)}, 
  title={Equivalent Lap Time Minimization Strategies for a Hybrid Electric Race Car}, 
  year={2018},
  volume={},
  number={},
  pages={6125-6131},
  doi={10.1109/CDC.2018.8618724}}

@article{optimal_tyre_management,
    title = {Optimal Tyre Management of a Formula One car},
    journal = {IFAC-PapersOnLine},
    volume = {53},
    number = {2},
    pages = {14456-14461},
    year = {2020},
    note = {21st IFAC World Congress},
    issn = {2405-8963},
    doi = {https://doi.org/10.1016/j.ifacol.2020.12.1446},
    author = {Wilhelm J. West and David J.N. Limebeer},
}

@article{optimizing_pit_stop_management_game_theory,
    title = {Optimizing pit stop strategies in Formula 1 with dynamic programming and game theory},
    journal = {European Journal of Operational Research},
    volume = {319},
    number = {3},
    pages = {908-919},
    year = {2024},
    issn = {0377-2217},
    doi = {https://doi.org/10.1016/j.ejor.2024.07.011},
    author = {Felipe Aguad and Charles Thraves}
}

@inproceedings{adaptation_algorithms_powertrain,
    title = "Adaptation algorithms for the hybrid electric powertrain of a race car",
    author = "Camillo Balerna and Mauro Salazar and Nicolas Lanzetti and Carlo Bussi and Christopher Onder",
    year = "2018",
    language = "English",
    booktitle = "FISITA World Automotive Congress",
    note = "37th FISITA World Automotive Congress 2018 ; Conference date: 02-10-2018 Through 05-10-2018"
}

@ARTICLE{learning_policies_automated_racing,
  author={Spielberg, Nathan A. and Templer, Maximilian and Subosits, John and Gerdes, J. Christian},
  journal={IEEE Open Journal of Intelligent Transportation Systems}, 
  title={Learning Policies for Automated Racing Using Vehicle Model Gradients}, 
  year={2023},
  volume={4},
  number={},
  pages={130-142},
  doi={10.1109/OJITS.2023.3237977}
}

@INPROCEEDINGS{energy_management_autonomous,
  author={Herrmann, Thomas and Christ, Fabian and Betz, Johannes and Lienkamp, Markus},
  booktitle={2019 IEEE Intelligent Transportation Systems Conference (ITSC)}, 
  title={Energy Management Strategy for an Autonomous Electric Racecar using Optimal Control}, 
  year={2019},
  volume={},
  number={},
  pages={720-725},
  doi={10.1109/ITSC.2019.8917154}
}

@INPROCEEDINGS{minimum_race_time_planning_strategy_autonomous,
  author={Herrmann, Thomas and Passigato, Francesco and Betz, Johannes and Lienkamp, Markus},
  booktitle={2020 IEEE 23rd International Conference on Intelligent Transportation Systems (ITSC)}, 
  title={Minimum Race-Time Planning-Strategy for an Autonomous Electric Racecar}, 
  year={2020},
  volume={},
  number={},
  pages={1-6},
  doi={10.1109/ITSC45102.2020.9294681}
}

@ARTICLE{learing_based_mpc_amz,
  author={Kabzan, Juraj and Hewing, Lukas and Liniger, Alexander and Zeilinger, Melanie N.},
  journal={IEEE Robotics and Automation Letters}, 
  title={Learning-Based Model Predictive Control for Autonomous Racing}, 
  year={2019},
  volume={4},
  number={4},
  pages={3363-3370},
  doi={10.1109/LRA.2019.2926677}
}

@article{farroni_temp,
author = {Flavio Farroni and Michele Russo and Aleksandr Sakhnevych and Francesco Timpone},
title ={TRT EVO: Advances in real-time thermodynamic tire modeling for vehicle dynamics simulations},

journal = {Proceedings of the Institution of Mechanical Engineers, Part D: Journal of Automobile Engineering},
volume = {233},
number = {1},
pages = {121-135},
year = {2019},
doi = {10.1177/0954407018808992},

}

@software{ceres,
  author = {Agarwal, Sameer and Mierle, Keir and The Ceres Solver Team},
  title = {{Ceres Solver}},
  license = {Apache-2.0},
  url = {https://github.com/ceres-solver/ceres-solver},
  version = {2.2},
  year = {2023},
  month = {10}
}

@book{pacejka,
  author    = {Hans B. Pacejka},
  title     = {Tire and Vehicle Dynamics},
  edition   = {3rd},
  year      = {2006},
  publisher = {Butterworth-Heinemann},
  address   = {Burlington, MA, USA}
}

@book{guiggiani,
  author    = {Massimo Guiggiani},
  title     = {The Science of Vehicle Dynamics: Handling, Braking, and Ride of Road and Race Cars},
  year      = {2022},
  publisher = {Springer International Publishing},
  address   = {Cham, Switzerland},
  isbn      = {978-3-031-06460-9}
}
	
\end{document}